\documentclass[conference]{IEEEtran}

\usepackage{amsmath,amssymb,amsfonts}
\usepackage{graphicx}
\usepackage{algorithm}
\usepackage{algpseudocode}
\usepackage{balance} 
\usepackage{booktabs} 
\usepackage{url}      
\usepackage{caption} 
\usepackage{subcaption} 
\usepackage{tikz}
\usetikzlibrary{shapes.geometric, arrows.meta, positioning}

\begin{document}

\title{Manifold-Aware Diffusion-Augmented Contrastive Learning for Noise-Robust Biosignal Representation}


\author{
\IEEEauthorblockN{Rami Zewail}
\IEEEauthorblockA{Computer Science \& Engineering Department\\
Egypt-Japan University of Science \& Technology (E-JUST)\\
New Borg El Arab, Alexandria, Egypt\\
Email: rami.zewail@ejust.edu.eg}
}

\maketitle

\begin{abstract}
Learning robust representations for physiological time-series signals continues to pose a substantial challenge in developing efficient few-shot learning applications. This is largely due to the complex  pathological variations  in biosignals. In this context, this paper introduces a manifold-aware Diffusion-Augmented Contrastive Learning (DACL) framework, which  efficiently  leverages the generative structure of latent diffusion models with the discriminative power of supervised contrastive learning. The proposed framework operates within  a contextualized scattering latent space  derived from Scattering Transformer (ST) features. Within a contrastive learning framework, we employ a forward diffusion  process in the scattering latent space as a structured manifold-aware feature augmentation technique. We assessed the proposed framework using the PhysioNet 2017 ECG benchmark dataset. The proposed method achieved a competitive AUROC of 0.9741 in the task of detecting atrial fibrillation from a single-lead ECG signal. The proposed framework achieved  performance on par with  relevant state-of-the-art related works. In-depth evaluation findings suggest that early stage diffusion serves as an ideal "local manifold explorer," producing embeddings with greater precision than typical augmentation methods while preserving inference efficiency.
\end{abstract}


\section{Introduction}
\IEEEPARstart{W}{ithin} the domain of biomedical artificial intelligence, the limited availability of expert-annotated data constitutes a significant impediment. In contrast to natural image domains, physiological signals such as ECG and EEG are costly to annotate and frequently exhibit significant class imbalances. This scarcity of data and expertise constrains the generalization potential of deep learning models.  Consequently, self-supervised learning frameworks are particularly attractive  in this area. 

The effectiveness of contrastive learning frameworks \cite{1, 2, 9} is fundamentally dependent on the quality of data augmentation employed to produce distinct yet semantically consistent "views" of a data sample. While the field of computer vision benefits from a well-established set of heuristic augmentations, the development of meaning-preserving transformations for physiological time-series data remains a significant and unresolved challenge \cite{3, 4, 11}. Simple transformations, such as jittering or permutation, often fail to respect the underlying physiological manifold, potentially compromising critical and pathological features. Therefore, there is a growing need for data-driven augmentation strategies rather than heuristic approaches.

Parallel to these developments, diffusion models \cite{6, 10} have emerged as powerful generative tools, exhibiting remarkable capabilities in modeling complex data distributions through progressive noising and denoising processes. While most research has focused on the \textit{reverse} process for high-fidelity generation, the forward noising process offers an overlooked but powerful property: it provides a mechanism for creating increasingly noisy iterations of a data point that remain mathematically grounded in the data's manifold structure \cite{7}.

Motivated by these concurrent advancements, we introduce \textbf{Diffusion-Augmented Contrastive Learning (DACL)}, a novel framework that substitutes heuristic augmentations with a structured learned strategy derived from the forward diffusion  process. Rather than synthesizing new samples through costly reverse sampling, DACL generates training views by sampling embeddings at varying noise levels along the forward diffusion trajectory. This methodology enables the encoder to learn representations that are invariant to manifold-consistent perturbations. To ensure that the input features are robust and structured, our framework operates within a contextualized latent space derived from the Scattering Transformer (ST) \cite{8}, adopting the computational efficiency paradigm of Latent Diffusion Models \cite{12}.

The specific contributions of this paper are as follows:
\begin{itemize}
    \item We propose \textbf{DACL}, an innovative hybrid framework that effectively integrates diffusion probabilistic models with supervised contrastive learning to derive noise-robust representations from biosignals. This framework operates within a context-aware scattering latent space derived using the Scattering Transformer ($\text{ST}$) \cite{8}.
    \item \textbf{DACL} introduces the application of the forward diffusion  process as a principled, learned augmentation strategy. This approach explicitly differentiates our method from generative synthesis techniques by circumventing the high computational demands of the reverse process while achieving superior manifold exploration.
    \item We establish that \textbf{DACL} serves as an effective "local manifold explorer" through its use of early-stage diffusion. Our framework achieves a patient-level AUROC of \textbf{0.9741} on the PhysioNet 2017 benchmark, substantially surpassing both standard contrastive baselines and reconstruction-based methods.
\end{itemize}

The rest of the paper is organized as follows: Section II provides a review of the related work in the literature. Section III presents the details of the proposed methodology. The experiments and results are presented in Section IV. A complete discussion is presented in Section V. Finally, Section VI concludes the paper.

\section{Related Work}

\begin{figure*}[!t]
    \centering
    \includegraphics[width=\textwidth]{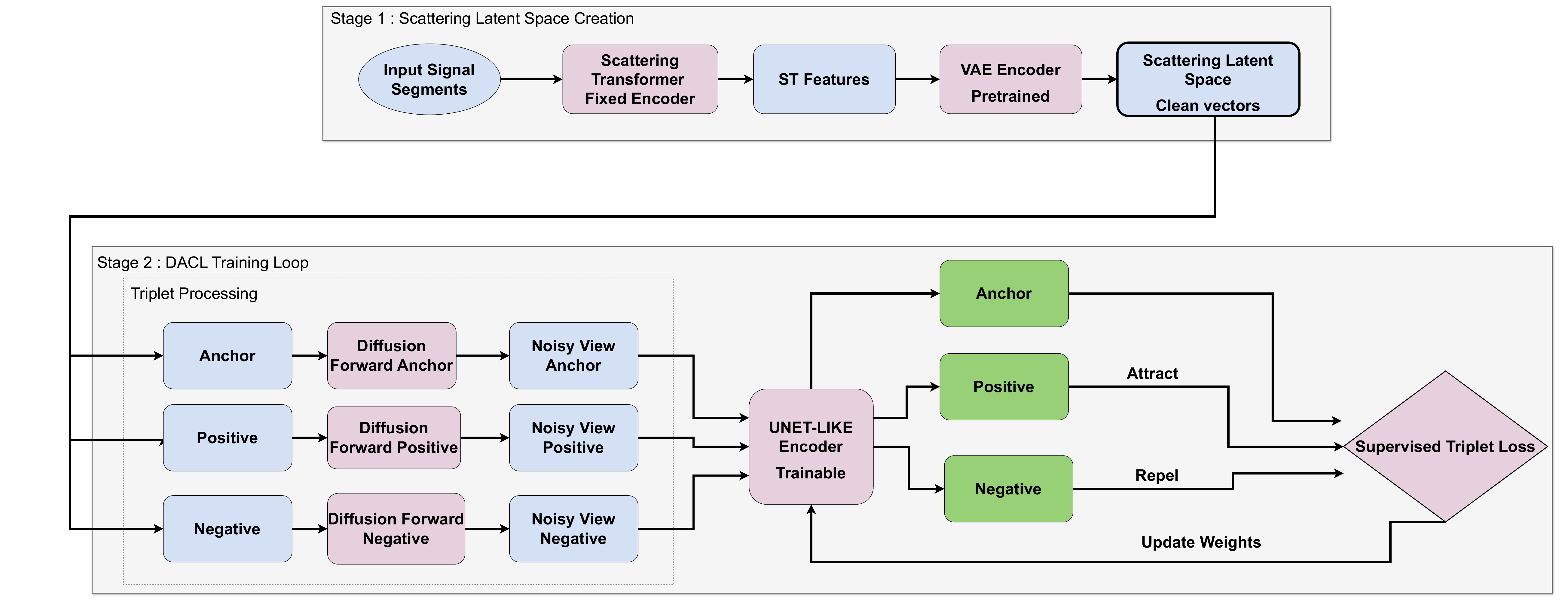}
    \caption{Conceptual diagram of the DACL framework. Stage I involves training a VAE on all ST features to create a general-purpose latent space. In Stage II, this latent space is used as the main training loop. A triplet of latent vectors is sampled, and noisy views are generated via the forward diffusion process. These views are passed through a shared U-Net encoder, normalized to the unit hypersphere, and trained via a supervised contrastive objective to produce noise-robust, class-discriminative embeddings.}
    \label{fig:figure1_workflow}
\end{figure*}

This section provides a comprehensive overview of the pertinent literature, emphasizing recent diffusion-based frameworks. We categorize diffusion-based methodologies into three distinct paradigms: \textit{Generative Data Augmentation}, \textit{Reconstruction-based Detection}, and \textit{Diffusion-Guided Representation Learning}.

\subsection{Generative Data Augmentation (Offline Synthesis)}
Diffusion Probabilistic Models (DPMs) have been used for the generation of high-quality artificial samples to mitigate data scarcity or class imbalances. For example, Yin et al. \cite{17} introduced the ``DiCL'' framework for bearing fault diagnosis, utilizing a fault-controllable DPM to generate diverse vibration signals for minority fault classes. In a similar physiological domain, Tan et al. introduced a Latent Diffusion Model for Heart Sound Synthesis \cite{23} to address challenges arising from highly imbalanced databases where normal heart sounds are more prevalent than abnormal ones. While effective, these methods function as offline augmentation strategies, incurring substantial computational costs due to the need to solve the reverse differential equation over numerous time steps ($T \approx 1000$ steps) to produce each new training sample. In addition, the downstream classifier is trained on synthetic data, which may introduce generative artifacts. In contrast, our proposed framework, DACL, employs a diffusion \textit{forward} process as an online feature augmentation strategy. Rather than synthesizing samples, we train the encoder to be invariant to manifold-consistent noise in a robust latent space, thereby achieving robustness without the inference overhead associated with generation.

\subsection{Reconstruction-based Anomaly Detection}
Recent studies have utilized the generative prior of Diffusion Probabilistic Models (DPMs) for anomaly detection through reconstruction. The underlying hypothesis is that a diffusion model, when trained solely on non-anomalous data, is incapable of accurately reconstructing anomalous regions. Diffusion-based methodologies have emerged as promising alternatives to conventional Generative Adversarial Networks (GANs) and Autoencoders. Wu et al. \cite{18} introduced a conditional diffusion model designed to reduce false alarms in intensive care unit (ICU) arrhythmia monitoring. Anomalies are detected by calculating the Mean Squared Error (MSE) between the predicted "healthy" waveform and the actual observed signal. Similarly, Zhang et al. \cite{19} proposed "ASD-Diffusion" for the detection of industrial acoustic anomalies, effectively utilizing the diffusion model as a high-capacity denoising autoencoder.

These methodologies are fundamentally \textit{reconstruction-based}, relying on sample-wise signal fidelity, which may be vulnerable to non-semantic variations (e.g., baseline wander or sensor noise) that are mathematically distinct yet clinically irrelevant to the task at hand. Furthermore, inference requires iterative denoising (with large diffusion time steps $T$) for reconstruction. In contrast, the DACL framework is \textit{discriminative}; it identifies anomalies based on distance within a learned metric space, using only a single encoder forward pass ($T=1$) during inference.

\subsection{Diffusion-Guided Representation Learning}
Our research falls within this emerging paradigm of employing diffusion processes for representation learning. Recent investigations have demonstrated that the denoising objective inherently promotes learning of semantic structures \cite{13, 14}. For instance, Li et al. \cite{15} introduced ContrastDM, which integrates contrastive learning with diffusion for hyperspectral classification, while Zeng et al. \cite{16} developed CLDF, which involves training a pixel decoder on features extracted from a frozen, pre-trained diffusion model. Our proposed framework, DACL, advances this paradigm by incorporating the diffusion forward process directly into the contrastive objective within an efficient, scattering-based latent space. In contrast to CLDF \cite{16}, which passively extracts features from a frozen network, DACL trains the encoder \textit{from scratch} against the diffusion trajectory, compelling the encoder to actively learn a manifold-invariant representation. Moreover, unlike deep clustering methods that rely solely on reconstruction (e.g., VAE-GAN \cite{20}), our proposed framework enforces both geometric compactness and discriminative separation in a signal processing-guided latent space.

\section{Methodology}

\subsection{Framework Overview}
This section presents the details of the proposed  \textbf{Diffusion-Augmented Contrastive Learning (DACL)} framework, a hybrid framework designed to learn noise-robust representations by leveraging principles from probabilistic diffusion models and supervised contrastive learning in a contextualized scattering latent space. The central tenet of our approach is the replacement of heuristic, handcrafted data augmentations with a principled stochastic transformation derived from the forward diffusion process, operating in an efficient signal processing-guided latent space. As illustrated in Fig. \ref{fig:figure1_workflow}, the framework operates in two distinct stages: (1) Manifold compression via a Variational Autoencoder (VAE) on Scattering Transformer (ST) features, and (2) Contrastive metric learning driven by diffusion-induced perturbations. The complete procedure for both training and inference is detailed in Algorithm \ref{alg:dacl}.

\subsection{Latent Space Construction} To create a seamless and compressed manifold conducive to diffusion operations, we initially extract fixed-size feature vectors utilizing the Scattering Transformer (ST) \cite{8}. This step yields contextualized, self-attentive, translation-invariant, and deformation-stable representations of raw biosignals. Subsequently, a lightweight Variational Autoencoder (VAE) is trained on these features to map the ST space $\mathcal{X}$ into a further compressed latent space $\mathcal{Z}$. This step is crucial because it ensures that the subsequent diffusion process operates on semantically meaningful latent variables rather than raw signal noise. For the downstream contrastive phase, we employ the mean of the VAE's learned posterior distribution, $\mathbf{z}_0 = \mu(\mathbf{x})$, as the "clean" anchor representation.

\subsection{Diffusion-Based Latent Augmentation}
Unlike generative data augmentation methods that require computationally expensive reverse sampling to synthesize new inputs \cite{17}, DACL leverages the forward diffusion  process as an efficient, on-the-fly augmentation strategy in a scattering latent space. For a given clean latent vector $\mathbf{z}_0$, we generate a ``noisy view'' $\mathbf{z}_t$ by sampling a time step $t$ and applying the closed-form Gaussian transition kernel:
\begin{equation}
    \mathbf{z}_t = \sqrt{\bar{\alpha}_t}\mathbf{z}_0 + \sqrt{1-\bar{\alpha}_t}\boldsymbol{\epsilon}, \quad \text{where } \boldsymbol{\epsilon} \sim \mathcal{N}(0, \mathbf{I}) \tag{1}
\end{equation}
Here, $\bar{\alpha}_t$ defines the noise schedule. This process injects structured, manifold-consistent noise into the representation, forcing the encoder to learn features that are invariant to the diffusion-induced perturbations. The time step $t$ acts as a control parameter for the augmentation intensity, allowing the model to explore the local neighborhood of the data manifold ($t \to 0$) or its global structure ($t \to T$).

\subsection{UNet-like Encoder Architecture}
The core discriminative model is a UNet-like encoder, denoted as $\text{Enc}_\theta$, which is adapted for vector-based input. While this architecture is structurally a conditional Multilayer Perceptron (MLP) composed of fully connected layers, it inherits the functional inputs of the standard diffusion noise predictor, processing the noisy latent vector $\mathbf{z}_{t}$ along with the corresponding time step $t$. This conditional structure enables the encoder to learn the time-dependent representations of the manifold. To ensure geometric stability within the metric space and mandate features that lie on the unit hypersphere, we strictly enforce L2 Normalization on the output embeddings. For an input view $\mathbf{z}_t$, the normalized embedding $\mathbf{e}$ is computed as:
\begin{equation}
\mathbf{e}_{\text{raw}}=\text{Enc}_{\theta}(\mathbf{z}_{t},t), \quad \mathbf{e}=\frac{\mathbf{e}_{\text{raw}}}{||\mathbf{e}_{\text{raw}}||_{2}+\xi} \tag{2}
\end{equation}
where $\xi$ is a small constant that ensures numerical stability. This projection ensures that all the embeddings lie on the unit hypersphere.

\subsection{Combined Contrastive Objective}
The network is optimized using a  loss function designed to simultaneously maximize inter-class separability and intra-class compactness. The total loss $L_{\text{Total}}$ is defined as:
\begin{equation}
    L_{\text{Total}} = L_{\text{Triplet}} + \lambda_{\text{Clustering}} L_{\text{KL-Clustering}} \tag{3}
\end{equation}
The primary component, Triplet Margin Loss, enforces discriminative distance constraints on the normalized embeddings as follows:
\begin{equation}
    L_{\text{Triplet}}(\theta) = \mathbb{E} \left[ \max(0, \|\mathbf{e}_A - \mathbf{e}_P\|_2^2 - \|\mathbf{e}_A - \mathbf{e}_N\|_2^2 + m) \right] \tag{4}
\end{equation}
where $(\mathbf{e}_A, \mathbf{e}_P, \mathbf{e}_N)$ represent the Anchor, Positive, and Negative embeddings. To further regularize the latent manifold, we incorporate the KL clustering loss. This term aligns the clean anchor embedding $\mathbf{e}_{\text{clean}}$ with a set of $K$ learnable cluster prototypes $\mathbf{C} = \{\mathbf{c}_k\}_{k=1}^K$, enforcing a tighter grouping for the dominant normal class and penalizing outliers.

\subsection{Inference Strategy} The proposed DACL framework offers a notable advantage in terms of inference efficiency. Unlike reconstruction-based anomaly detection methods, which require iterative reverse diffusion during inference ($T \approx 50-1000$ steps) \cite{18}, our approach requires only a single forward pass. For a given test sample $\mathbf{x}_{\text{test}}$, it is mapped into the latent space $\mathbf{z}_{0, \text{test}}$ and processed through the encoder in a single forward pass\textbf{ ($t=1$)}. The anomaly score is then computed efficiently as the Euclidean distance to the pre-computed normal prototype $\boldsymbol{\mu}_{\text{normal}}$ within the normalized embedding space, as follows: \begin{equation}\text{Score} = \|\mathbf{e}_{\text{test}} - \boldsymbol{\mu}_{\text{normal}}\|_2 \tag{5}\end{equation} This strategy effectively leverages the model's learned sensitivity to manifold perturbations while maintaining the speed characteristic of a standard discriminative classifier.

\begin{algorithm}[t!]
\caption{DACL Training and Inference Procedure}
\label{alg:dacl}
\small 
\begin{algorithmic}[1]
\State \textbf{Require:} Training features $\mathcal{X}$, Labels $\mathcal{Y}$, Stability constant $\xi = 1\text{e-}6$.
\State \textbf{Phase 1: Manifold Compression}
\State Train VAE on $\mathcal{X}$;
\State Extract latent set $\mathcal{Z}_0 \leftarrow E_{VAE}(\mathcal{X})$.
\State Initialize cluster prototypes $\mathbf{C} = \{\mathbf{c}_k\}_{k=1}^K$ and Encoder $\theta$.
\State \textbf{Phase 2: Diffusion-Augmented Training}
\For{each training batch}
    \State Sample triplet $(\mathbf{z}_{A}, \mathbf{z}_{P}, \mathbf{z}_{N})$ from $\mathcal{Z}_0$.
    \State Sample timesteps $t_A, t_P, t_N \sim \mathcal{U}(1, T_{train})$.
    \For{$v \in \{A, P, N\}$} \Comment{Process Augmentation}
        \State Generate noisy view $\mathbf{z}_{t_v}$ via Eq. (1).
        \State $\mathbf{e}_{\text{raw}, v} \leftarrow \text{Enc}_\theta(\mathbf{z}_{t_v}, t_v)$
        \State $\mathbf{e}_v \leftarrow \mathbf{e}_{\text{raw}, v} / (\|\mathbf{e}_{\text{raw}, v}\|_2 + \xi)$ \Comment{L2 Norm with stability term}
    \EndFor
    \State Update $\theta, \mathbf{C}$ minimizing $L_{\text{Total}}$ (Eq. 3).
\EndFor
\State \textbf{Phase 3: Inference (Anomaly Detection)}
\State Given test $\mathbf{x}_{\text{test}}$, project $\mathbf{z}_{0, \text{test}} \leftarrow E_{VAE}(\mathbf{x}_{\text{test}})$.
\State $\mathbf{e}_{\text{test}} \leftarrow \text{Enc}_\theta(\mathbf{z}_{0, \text{test}}, t=1)$, normalized using $\xi$. \Comment{Single forward pass at t=1}
\State \textbf{return} Score $\|\mathbf{e}_{\text{test}} - \boldsymbol{\mu}_{\text{normal}}\|_2$.
\end{algorithmic}
\end{algorithm}

\subsection{Dataset and Preprocessing}
The proposed framework was evaluated using the PhysioNet 2017 Challenge Dataset \cite{5}, a widely recognized benchmark for ECG rhythm classification. The dataset consists of single-lead ECG recordings categorized into \textit{Normal}, \textit{Atrial Fibrillation} (AF), and \textit{Other} rhythms. To frame the problem as a rigorous anomaly detection task, we designated the \textit{Normal} recordings as the in-distribution (healthy) class, while the \textit{AF} and \textit{Other} rhythms were combined to form a single \textit{Anomaly} class. To prevent data leakage and ensure subject independence, the dataset was partitioned by patient ID, with 70\% of patients allocated for training and 30\% for testing.

\subsection{Evaluation Protocol and Baselines}
Performance was reported using the patient-level Area Under the Receiver Operating Characteristic (AUROC), alongside Precision, Recall, and F1-Score. To isolate the contribution of diffusion-based augmentation in the scattering latent space, we compared DACL with three rigorous baselines. To ensure a fair comparison, all contrastive baselines were implemented using the unit hypersphere projection mechanism.

\begin{enumerate}
    \item \textbf{Contrastive + Gaussian Aug. (Baseline 1):} This represents the standard heuristic approach. It is a supervised contrastive model in which ``views'' are generated by adding isotropic Gaussian noise ($\epsilon \sim \mathcal{N}(0, \sigma^2 \mathbf{I})$) to the latent vectors. This directly tests whether \textit{structured} diffusion noise is superior to \textit{random} noise.
    \item \textbf{Denoising Autoencoder (DAE) (Baseline 2):} This represents the reconstruction-based paradigm. A standard DAE was trained to minimize the Mean Squared Error (MSE) between the noisy and clean inputs. This tests whether \textit{discriminative distance} is superior to \textit{sample reconstruction}.
    \item \textbf{DACL (Triplet Only):} An ablation of our framework that utilizes diffusion augmentation but removes the $\lambda_{\text{Clustering}}$ term. This specifically isolates the effect of the deep-clustering objective.
\end{enumerate}

\subsection{Internal Benchmarking Results}
All models were optimized using the Adam optimizer with a learning rate of $1\text{e-}4$ and a batch size of 64. A critical component of the DACL framework is the 32-dimensional scattering latent space. As detailed in Table \ref{tab:main_results}, DACL (Triplet+Clustering) achieved the \textbf{highest patient-level AUROC of 0.9741 among all comparison baselines.}

The comparison with the Gaussian baseline is particularly revealing. Whereas Gaussian noise yields a high recall (0.9005), it suffers from significantly lower precision (0.5408). This indicates that the Gaussian model exhibits a bias toward hypersensitivity, erroneously flagging many healthy samples as anomalous. In contrast, DACL achieves a superior balance, boosting the precision to 0.7464 while maintaining high recall. This confirms that structured, diffusion-induced perturbations assist the encoder in defining a tighter and more precise decision boundary around the normal manifold.

Furthermore, comparing the full DACL framework with the triplet-only variant highlights the necessity of the deep clustering objective. Although the triplet-only configuration achieved a competitive AUROC (0.9508), its low precision (0.4821) and consequently reduced F1-Score (0.6313) indicate an insufficient consolidation of the normal class manifold. Therefore, the integration of the KL-Clustering term is essential for enforcing geometric compactness, serving as the decisive factor in achieving an optimal balance between sensitivity and specificity.

\begin{table}[h!]
\centering
\caption{Performance comparison on PhysioNet 2017. DACL achieves the optimal trade-off between sensitivity and specificity, significantly outperforming the reconstruction-based DAE and the heuristic Gaussian baseline.}
\label{tab:main_results}
\resizebox{\columnwidth}{!}{%
\begin{tabular}{@{}lccccc@{}}
\toprule
\textbf{Model} & \textbf{AUROC} & \textbf{Precision} & \textbf{Recall} & \textbf{F1-Score} & \textbf{Accuracy} \\ \midrule
\textbf{DACL (Full)} & \textbf{0.9741} & \textbf{0.7464} & \textbf{0.9321} & \textbf{0.8290} & \textbf{0.9544} \\
DACL (Triplet Only) & 0.9508 & 0.4821 & 0.9140 & 0.6313 & 0.8638 \\
Gaussian Aug. & 0.9513 & 0.5408 & 0.9005 & 0.6757 & 0.8898 \\
Denoising AE & 0.7675 & 0.2767 & 0.6561 & 0.3893 & 0.7374 \\ \bottomrule
\end{tabular}%
}
\end{table}

\begin{figure}[!t]
    \centering
    \includegraphics[width=3.4in]{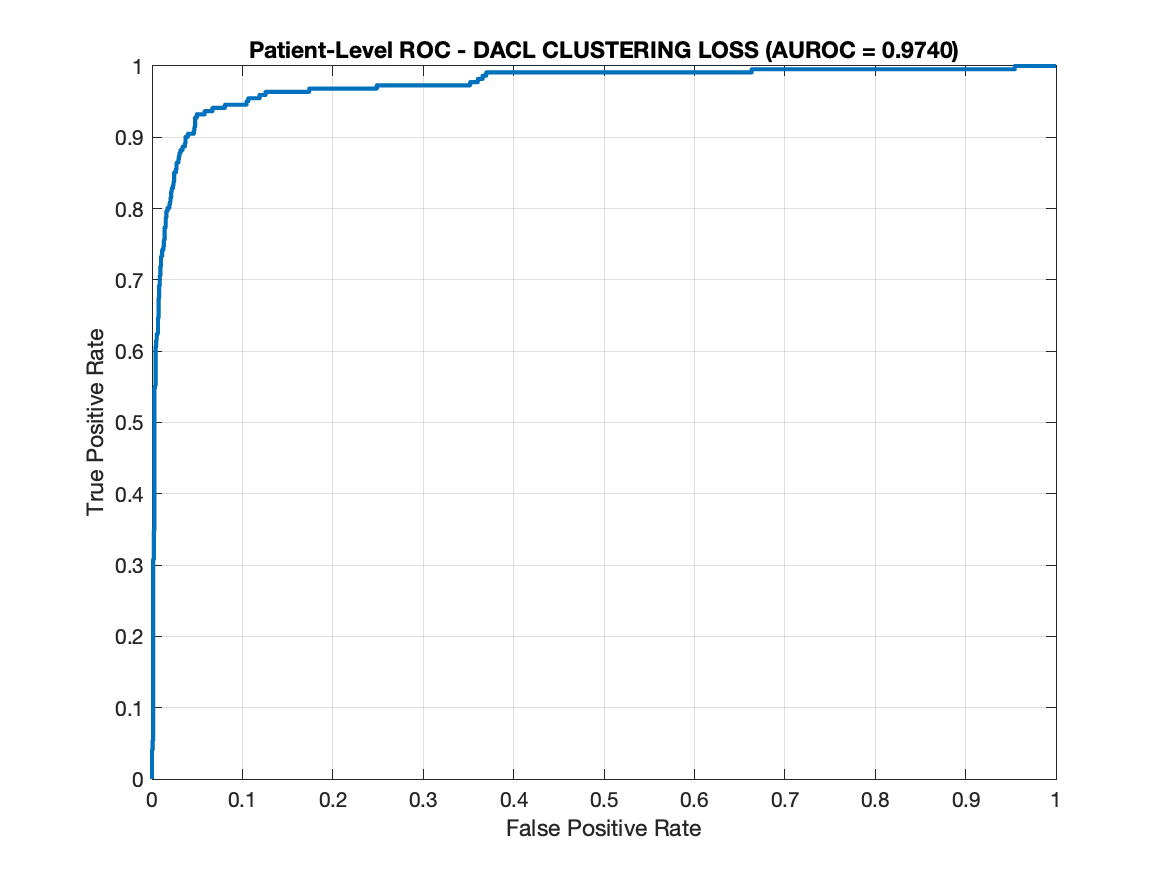} 
    \caption{Patient-Level ROC Curve for the DACL model utilizing the Combined Contrastive Loss on PhysioNet 2017. The curve demonstrates robust performance with an AUROC of 0.9741, confirming high sensitivity at low false-positive rates.}
    \label{fig:figure2_roc}
\end{figure}

\begin{figure}[!t]
    \centering
    \includegraphics[width=3.4in]{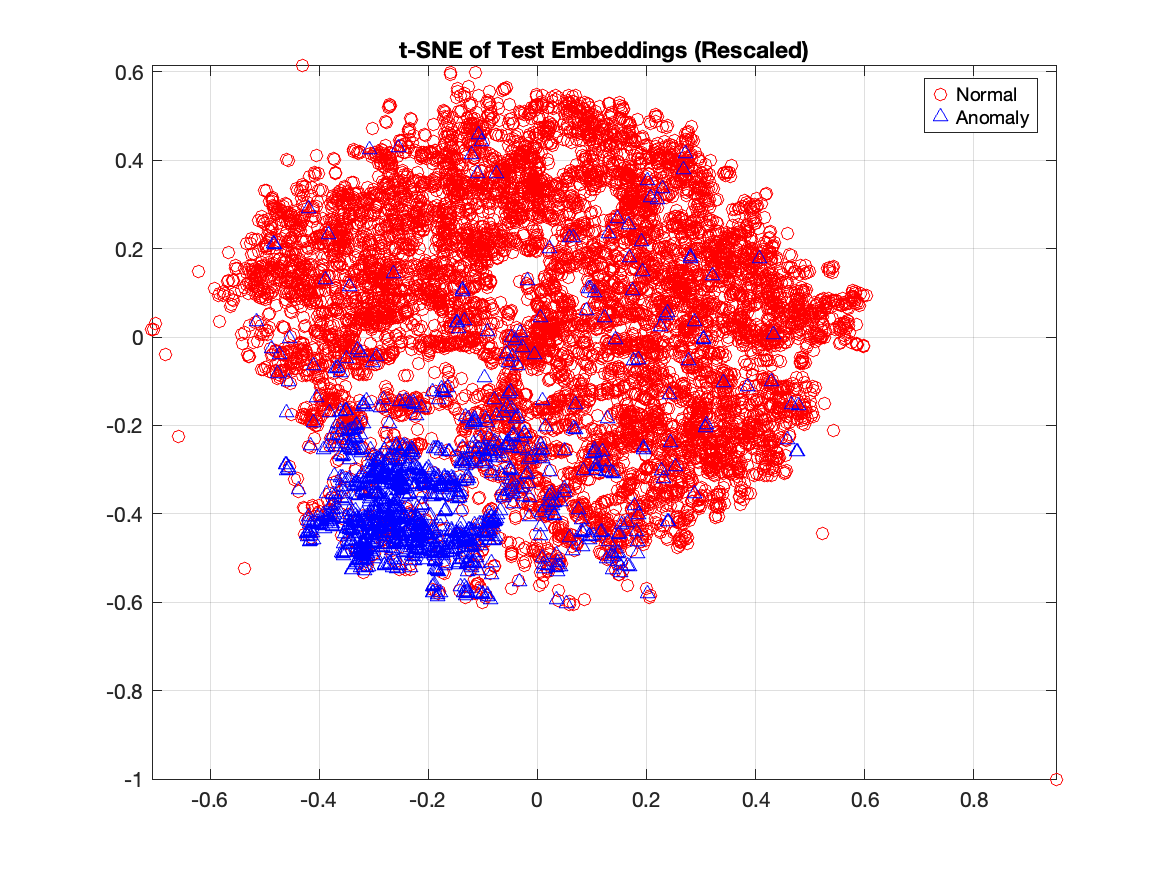} 
    \caption{t-SNE visualization of the test embeddings generated by the DACL encoder. The clusters show markedly improved separation compared to baselines, with the Normal class (Red) forming a compact, well-defined manifold distinct from anomalies.}
    \label{fig:figure3_tsne}
\end{figure}

\subsection{Comparison with State-of-the-Art Frameworks}
To thoroughly assess the proposed framework (\textbf{DACL}), we benchmarked it against established external baselines, categorizing them according to their learning paradigms (Table \ref{tab:sota_quantitative}).

\subsubsection{Self-supervised vs. Supervised SOTA}
A fundamental objective of representation learning is to achieve parity with the performance of fully supervised models. Notably, DACL (0.9741 AUROC) substantially surpasses the attention-based supervised model \textbf{MINA} \cite{21} (0.9488) and effectively aligns with the performance of the state-of-the-art supervised \textbf{CWT-MB-ResNet} \cite{22} (0.9761). This finding illustrates that manifold-aware diffusion augmentation enables the encoder to capture highly discriminative pathological features, thereby competing effectively with supervised deep learning approaches.

\subsubsection{Efficiency vs. Diffusion Baselines}
The proposed framework (DACL) introduces an innovative application of the diffusion paradigm by leveraging its capabilities exclusively for representation learning within an efficient scattering latent space while preserving the simplicity of single-pass inference. Although recent diffusion-based methods, such as Wu et al. \cite{18} (reconstruction) and \textbf{DiCL} \cite{17} (Synthetic Data), demonstrate potential, they are hindered by high inference latency due to the need for iterative reverse sampling ($T \approx 200-1000$). By employing the diffusion forward process for metric learning rather than for generation or reconstruction, the DACL framework operates with a single encoder forward pass ($T=1$). This results in a speedup factor of approximately $T$, making it particularly suitable for real-time clinical monitoring, where low latency is essential.

\begin{table*}[t!]
\centering
\caption{Quantitative Comparison. \textbf{DACL} matches the performance of fully supervised State-of-the-Art methods (MINA, CWT-MB-ResNet) while operating as a self-supervised anomaly detector. Furthermore, it maintains an inference cost of $T=1$, offering a significant speedup over reconstruction-based diffusion methods which require $T \gg 1$.}
\label{tab:sota_quantitative}
\begin{tabular}{@{}llllccc@{}}
\toprule
\textbf{Method} & \textbf{Paradigm} & \textbf{Target Domain} & \textbf{Inference Mechanism} & \textbf{Cost ($T$)} & \textbf{AUROC} & \textbf{Acc / F1} \\ \midrule
\multicolumn{7}{c}{\textit{Independent Supervised SOTA (PhysioNet 2017)}} \\ \midrule
\textbf{MINA [21]} & Supervised (Attn) & ECG & Attention Network & 1 & 0.9488 & - \\
\textbf{CWT-MB-ResNet [22]} & Supervised (CNN) & Scalogram & Multi-Branch ResNet & 1 & 0.9761 & - \\ \midrule
\multicolumn{7}{c}{\textit{Direct Comparison (This Work)}} \\ \midrule
\textbf{Gaussian Baseline} & Contrastive & ECG & Encoder Forward Pass & 1 & 0.9513 & 88.98\% (Acc) \\
\textbf{Denoising AE} & Reconstruction & ECG & Encoder-Decoder & 1 & 0.7675 & - \\
\textbf{DACL (Ours)} & \textbf{Latent Aug.} & \textbf{ECG} & \textbf{Encoder Forward Pass} & \textbf{1} & \textbf{0.9741} & \textbf{95.44\% (Acc)} \\ \midrule
\multicolumn{7}{c}{\textit{External Reference }} \\ \midrule
\textbf{Wu et al. [18]} & Diffusion-Reconstruction & ICU Vitals & Iterative Denoising & $\approx 200$ & - & 91.71\% (Acc) \\
\textbf{DiCL [17]} & Diffusion-Synthesis & Vibration & Reverse Sampling & $\approx 1000$ & - & 99.82\% (Acc) \\ 
\textbf{HeartSound-LDM [23]} & Diffusion-Synthesis & Heart Sounds & Reverse Sampling & $\approx 1000$ & - & 93.4\% Acc) \\
\textbf{TS-TCC [4]} & Heuristic Contrastive & EEG & Encoder Forward Pass & 1 & - & 83.0\% (Acc) \\ \bottomrule
\end{tabular}
\vspace{1ex}
\end{table*}

\subsection{Ablation Study: Manifold Exploration}
\begin{figure}[!t]
    \centering
    \includegraphics[width=3.4in]{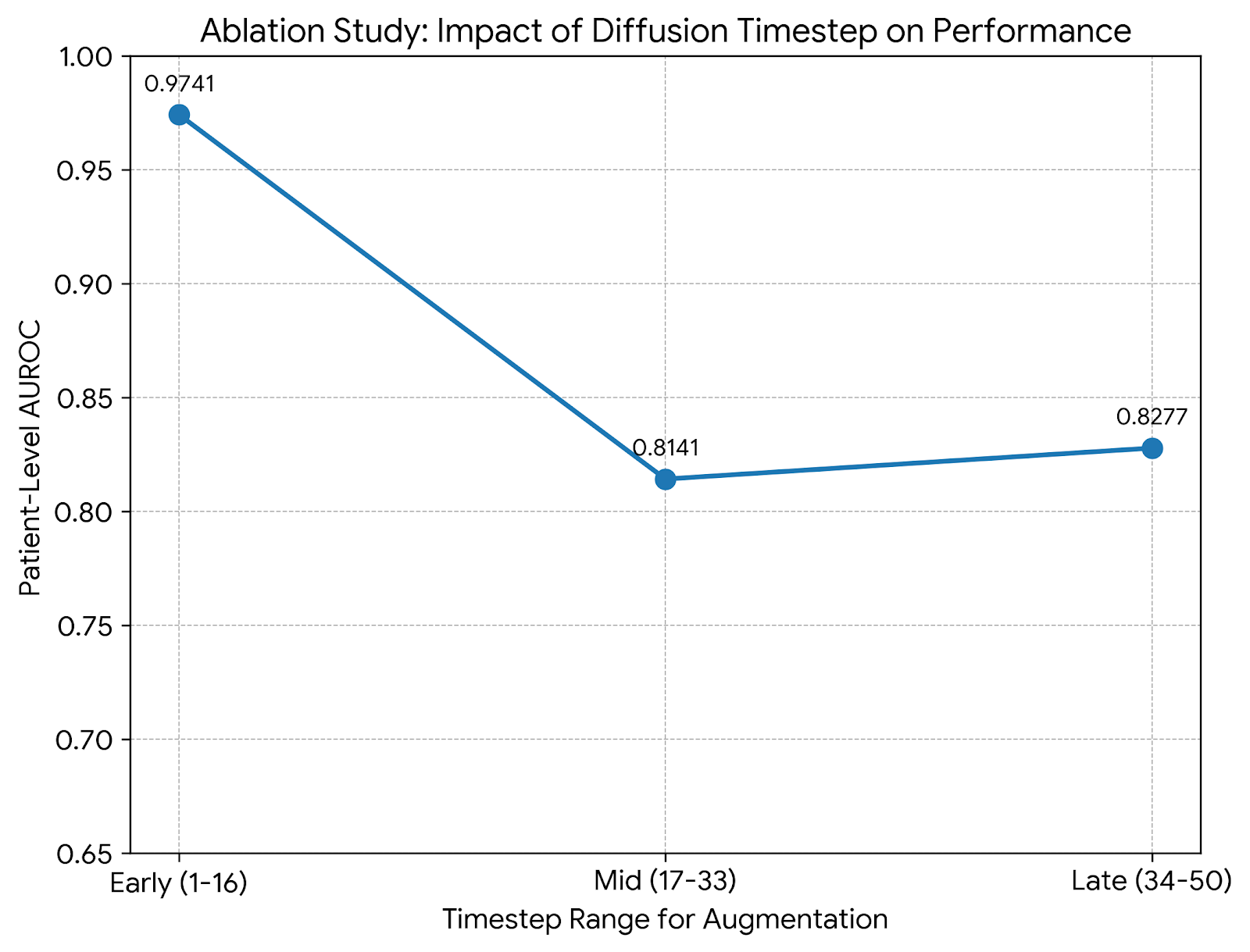}
    \caption{Results from the diffusion timestep ablation study. Performance clearly peaks during Early timesteps ($t \le 16$) and degrades as noise intensity increases, confirming that local manifold exploration is superior to heavy corruption for this task.}
    \label{fig:figure4_ablation}
\end{figure}

To examine the influence of noise intensity on the representation quality, we trained the DACL framework using restricted diffusion time step ranges (Fig. \ref{fig:figure4_ablation}). The findings indicate a distinct optimal range for manifold exploration in the study. Performance reaches its peak during the early stage diffusion ($t \in [1, 16]$), achieving a maximum AUROC of 0.9741. This suggests that minor-structured perturbations enable the encoder to effectively explore the local neighborhood of the scattering manifold without compromising its semantic identity. Conversely, performance deteriorates markedly during the Mid ($t \in [17, 33]$) and Late ($t \in [34, 50]$) stages. This observation confirms that excessive diffusion noise ultimately degrades the fine-grained pathological features essential for anomaly detection, thereby displacing the data from the \textbf{semantic scattering manifold}.

\section{Discussion}

This study introduces DACL, a framework that utilizes the forward diffusion  process within a scattering latent space as a systematic and learnable augmentation strategy for contrastive physiological representation learning. The empirical success of this method, demonstrated by an outstanding AUROC of \textbf{0.9741}, provides several critical insights into the nature of self-supervised learning for biosignals.

\subsection{Precision and Manifold Topology}
The pronounced performance disparity between DACL and the Gaussian Baseline underscores a fundamental limitation inherent in heuristic augmentations. Notably, the Gaussian model's commendable AUROC (0.9513) confirms the inherent robustness of the underlying scattering latent space. However, its low precision (0.5408) suggests a ``hypersensitive'' decision boundary. The introduction of isotropic noise results in the Gaussian model uniformly expanding the rejection zone in all directions, thereby erroneously identifying valid but rare physiological variations as anomalies. In contrast, DACL achieved markedly higher precision (0.7464). This improvement is attributed to the \textbf{manifold-aware} characteristics of the diffusion noise. Given that the diffusion kernel is conditioned on the data distribution, the generated ``views'' remain semantically valid. Consequently, the encoder can delineate a decision boundary that precisely aligns with the true physiological manifold, effectively rejecting anomalies without misclassifying healthy edge cases.

\subsection{Impact of the KL-Clustering Objective}
The ablation study indicated that the triplet-only loss function encountered difficulties in achieving high precision ($\approx 0.48$). The incorporation of the KL-clustering loss was crucial for stabilizing the metric space. Even with the unit hypersphere constraint applied, the triplet loss separates negative samples but does not necessarily ensure compactness within the normal class. The clustering objective serves as a significant geometric regularizer, compelling normal data to converge into a high-density region centered on the prototype $\boldsymbol{\mu}_{\text{normal}}$. This density convergence enhances the reliability of the Euclidean distance metric utilized for inference.

\subsection{Discriminative Distance vs. Sample-Level Reconstruction}
The suboptimal performance of the Denoising Autoencoder (AUROC 0.7675) underscores a significant observation: \textit{sample-level reconstruction is often inadequate for semantic anomaly detection in complex biosignals.} Standard DAEs prioritize low-level signal fidelity and frequently expend capacity reconstructing non-essential high-frequency noise or baseline wander. Conversely, the proposed DACL framework learns \textit{discriminative distances} within a scattering latent space by optimizing feature invariance rather than signal reconstruction. Consequently, it is more adept at disregarding clinically irrelevant artifacts and concentrating exclusively on the structural deviations that characterize arrhythmia.

\section{Conclusion}
This research presented \textbf{Diffusion-Augmented Contrastive Learning (DACL)}, a pioneering approach that successfully merges generative diffusion models with discriminative representation learning. The framework leverages the forward diffusion  process as a structured, manifold-consistent augmentation technique in a context-aware scattering latent space.

Our empirical assessment of the PhysioNet 2017 benchmark reveals that DACL achieves a detection performance on par with leading supervised models (AUROC \textbf{0.9741}) while operating as a self-supervised anomaly detector. In contrast to generative baselines that demand computationally intensive reverse sampling ($T \gg 1$), DACL functions with a single encoder forward pass, making it exceptionally suitable for real-time clinical applications. Our examination of the noise trajectory demonstrates that \textbf{early-stage diffusion} acts as an optimal local manifold explorer, bolstering model robustness without sacrificing semantic integrity. These results underscore the promising potential of signal processing-guided latent diffusion models in biomedical AI.

\balance

\end{document}